\documentclass{article}

\usepackage[preprint]{neurips_2025}

\usepackage[utf8]{inputenc} 
\usepackage[T1]{fontenc}    
\usepackage{hyperref}       
\usepackage{url}            
\usepackage{booktabs}       
\usepackage{amsfonts}       
\usepackage{nicefrac}       
\usepackage{microtype}      
\usepackage{xcolor}         
\usepackage{multirow}       
\usepackage{adjustbox}
\usepackage{graphicx}
\usepackage{makecell}
\usepackage{amsmath}
\usepackage{float}
\usepackage{subcaption}
\usepackage[most]{tcolorbox}
\usepackage{natbib}
\setcitestyle{authoryear,open={(},close={)}}

\title{Uncertainty Profiles for LLMs: Uncertainty Source Decomposition and Adaptive Model-Metric Selection}

\author{%
  Pei-Fu Guo \\
  National Taiwan University\\
  \text{r12922217@csie.ntu.edu.tw} \\
  \And
  Yun-Da Tsai \\
  National Taiwan University\\
  \text{f08946007@csie.ntu.edu.tw} \\
  \AND
  Shou-De Lin \\
  National Taiwan University\\
  \text{sdlin@csie.ntu.edu.tw} \\
}

\begin{document}

\maketitle

\begin{abstract} 
  Large language models (LLMs) often generate fluent but factually incorrect outputs, known as hallucinations, which undermine their reliability in real-world applications. While uncertainty estimation has emerged as a promising strategy for detecting such errors, current metrics offer limited interpretability and lack clarity about the types of uncertainty they capture. In this paper, we present a systematic framework for decomposing LLM uncertainty into four distinct sources, inspired by previous research. We develop a source-specific estimation pipeline to quantify these uncertainty types and evaluate how existing metrics relate to each source across tasks and models. Our results show that metrics, task, and model exhibit systematic variation in uncertainty characteristic. Building on this, we propose a method for task specific metric/model selection guided by the alignment or divergence between their uncertainty characteristics and that of a given task. Our experiments across datasets and models demonstrate that our uncertainty-aware selection strategy consistently outperforms baseline strategies, helping us select appropriate models or uncertainty metrics, and contributing to more reliable and efficient deployment in uncertainty estimation. Code is available at \href{https://anonymous.4open.science/r/llm-uncertainty-profile}{link}.
\end{abstract}

\section{Introduction}
\label{sec:intro}
Large language models (LLMs) have demonstrated remarkable capabilities in natural language generation, often surpassing average human performance on tasks involving mathematics, reasoning, and programming. Despite these advances, LLMs frequently produce confident, yet factually incorrect responses, commonly referred to as \textit{hallucinations}. Such errors pose a serious challenge to the reliability and trustworthiness of LLM outputs.

A promising direction for addressing hallucinations lies in \textit{LLM uncertainty estimation}. Previous research has shown that hallucinations can often be identified by examining how uncertain LLM is when responding to a prompt. This has led to the development of a variety of uncertainty estimation techniques. Broadly, these methods can be grouped into four categories: likelihood-based approaches \citet{malinin2020uncertainty, kuhn2023semantic}, consistency/ensemble-based methods \citet{xiong2023can, jiang2023calibrating}, verbalization-based techniques \citet{kadavath2022language, lin2022teaching}, and approaches specifically targeting aleatoric and epistemic uncertainty \citet{ahdritz2024distinguishing, yadkori2024believe, gao2024spuq}.

Although these methods have shown strong predictive performance across different datasets and model families, they suffer from two key limitations. First, most uncertainty metrics do not provide actionable information on why a particular response is uncertain. As a result, users receive little guidance on how to reduce uncertainty or improve input accuracy. Second, it remains unclear which specific sources of uncertainty each metric captures, such as linguistic ambiguity, knowledge gaps, or the inherent randomness of the decoding process. This lack of clarity makes it difficult for practitioners to select the most appropriate metrics for new tasks. Instead, users often rely on prior benchmark results or choose metrics based on intuitive interpretations of their design, rather than a systematic understanding of what the metrics actually capture. These gaps highlight the need for a systematic method to analyze the sources of uncertainty in LLM-generated responses, enabling better interpretation, selection, and application of uncertainty estimation in practice.

In this paper, we aim to answer the following key questions:
\begin{enumerate}
    \item What type of uncertainty source does existing uncertainty metric primarily capture?
    \item Do certain tasks consistently exhibit stronger tendencies toward specific uncertainty source?
    \item Are different language models more susceptible to specific uncertainty source?
    \item Can we leverage uncertainty characteristic to adaptively select appropriate metrics and models for different downstream task?
\end{enumerate}

To address these questions, we begin by identifying four distinct sources of uncertainty in LLM responses, motivated by prior work on improving model response accuracy. We then develop a pipeline with dedicated estimators to quantify each uncertainty source. By analyzing the relation between source estimators and existing LLM uncertainty metrics, we find that different metrics are sensitive to different types of uncertainty. Our analysis also shows that tasks tend to exhibit distinct uncertainty profiles, and that models vary in their susceptibility to different uncertainty sources. Building on these findings, we propose an adaptive approach for selecting uncertainty metrics and models based on the estimated uncertainty profile of a task. Our experimental results show that our method consistently outperforms non-adaptive baselines, offering a practical alternative to manual tuning or trial-and-error strategies for efficient deployment in real-world uncertainty estimation applications.

The major contributions of this paper are as follows:
\begin{itemize}
    \item We propose a conceptual framework that decomposes LLM uncertainty into four interpretable sources, represented as an \textit{uncertainty profile}—a vector capturing the fine-grained structure of model uncertainty.
    \item We develop an estimation pipeline that quantifies each uncertainty source at the response level and analyze these uncertainty profiles across multiple metrics, tasks, and models.
    \item We present a novel method for adaptive model and metric selection based on uncertainty profiles, enabling more effective and reliable deployment in real-world uncertainty estimation scenarios.
\end{itemize}

\nocite{tsai2023differential}
\nocite{da2022fast}
\nocite{guo2023towards}
\nocite{tsai2024code}
\nocite{tsai2024text}
\nocite{tsai2023automl}
\nocite{liu2024craftrtl}
\nocite{tsai2024lil}
\nocite{guo2024benchmarking}
\nocite{wu2024linearapt}
\nocite{tsai2024handling}
\nocite{tsai2023rtlfixer}
\nocite{tsai2024enhance}
\nocite{yen2024enhance}
\nocite{tsai2024toward}
\nocite{liawi2023psgtext}

\section{Profiling LLM Uncertainty: A Literature-Grounded Taxonomy}
\label{sec:uncertainty_source_type}
Uncertainty in LLM responses can arise from a wide range of factors, many of which are entangled and difficult to isolate in practice. To better understand and interpret what existing uncertainty metrics are capturing, we propose a decomposition of LLM uncertainty into four principal sources. This taxonomy is developed through a broad survey of prior work on model calibration, uncertainty estimation, and failure analysis in language models. While not exhaustive, these four sources reflect distinct and recurring behavior patterns in LLM outputs that we believe are both interpretable and practically useful.
We introduce the four principal uncertainty sources as follows:

\textbf{Surface Form Uncertainty (SU)} arises when the model struggles to parse the linguistic structure of the input, such as rare vocabulary, complex syntax, or unconventional phrasing. This often leads to shallow comprehension failures. For instance, \citet{zhou2024paraphrase} show that even subtle changes in surface form can significantly affect answer distributions and success rates in math reasoning tasks, highlighting the sensitivity of LLMs to low-level linguistic variation.

\textbf{Aleatoric Uncertainty (AU)} reflects inherent ambiguity or underspecification in the prompt. Even when syntactically clear, the input may support multiple valid interpretations, leading to diverse or inconsistent responses. This source of uncertainty is particularly important in instruction-following or open-ended tasks. \citet{hou2023decomposing} decompose total uncertainty in LLM responses into aleatoric and epistemic components using input clarification ensembling. \citet{gao2024spuq} jointly handle aleatoric and epistemic uncertainty by proposing a perturbation and aggregation module.

\textbf{Epistemic Uncertainty (EU)} reflects the model’s limited factual knowledge or reasoning capacity. In such cases, the model lacks the necessary information or logical grounding to produce a correct response. Prior work by \citet{yadkori2024believe}, \citet{ling2024uncertainty} and \citet{ahdritz2024distinguishing} specifically focuses on isolating and estimating epistemic uncertainty in LLMs, emphasizing its importance for evaluating model reliability. 

\textbf{Operational Uncertainty (OU)} refers to errors that emerge during the response generation process. These include stochastic decoding issues such as logical slips, arithmetic mistakes, or formatting inconsistencies. Studies like \citet{bigelow2024forking} demonstrate that small perturbations at critical decoding steps can lead to dramatically different outputs, while \citet{xu2025cod} and \citet{zeng2025revisiting} show that longer reasoning chains tend to amplify such instabilities—underscoring the impact of decoding dynamics on response quality.

While not exhaustive, these four sources capture a diverse and practically important subset of the uncertainty phenomena commonly observed in LLMs. By isolating and analyzing these specific modes of failure, we aim to provide a more structured understanding of model behavior—one that can inform the development of more targeted evaluation metrics and more effective mitigation strategies.

\section{Uncertainty Source Estimation and Validation}
\label{sec:uncertainty_source_estimation}
In this section, we first propose a novel, structured framework for estimating the four uncertainty sources introduced in Section~\ref{sec:uncertainty_source_type}. Then we validate the predictability of our estimator to demonstrate the effectiveness of our framework. 

\begin{figure}
\vspace{-1.3cm}
    \centering
    \includegraphics[width=0.9\linewidth]{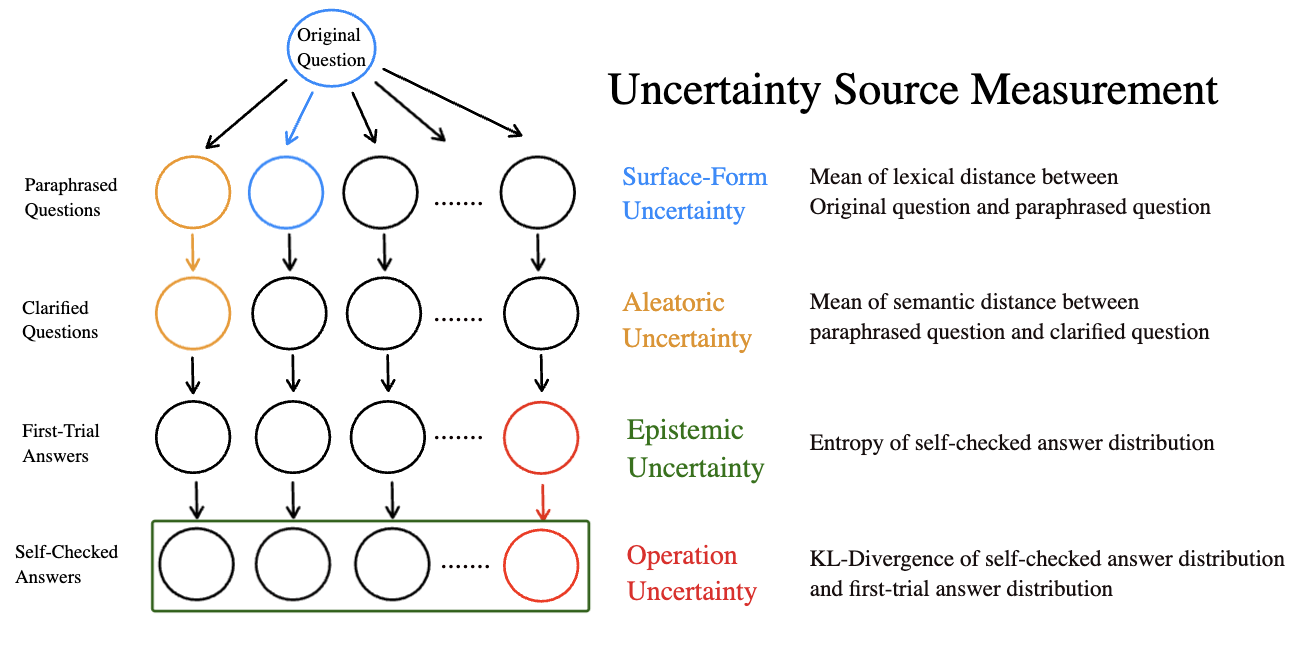}
    \caption{\textbf{Uncertainty Source Estimation Pipeline.} 
Our framework decomposes LLM uncertainty into four interpretable sources through a structured, multi-stage prompting process. Starting from a single question, the model generates multiple response chains, each progressing through paraphrasing, clarification, answering, and self-checking stages. Differences observed at each stage are used to estimate the corresponding source of uncertainty.}

    \label{fig:source_tree}
\end{figure}

\subsection{Estimation Pipeline}
Building on the conceptual foundations of the four uncertainty sources, we design a multi-stage prompting framework that estimates each source by isolating key steps in the question-answering process. As illustrated in Figure~\ref{fig:source_tree}, the method starts from a single original question and generates multiple independent response chains. Each chain progresses through four structured stages, corresponding to different aspects of uncertainty highlighted in prior work.

\paragraph{Paraphrasing:} Starting from the original question, we generate multiple paraphrased variants using the model itself. These variants differ in surface form but aim to preserve the underlying meaning. By comparing the lexical differences between the original and the paraphrases, we can assess how sensitive the model is to superficial variations, capturing SU.
    
\paragraph{Clarification:} For each paraphrased question, we prompt the model to elaborate or clarify the question by adding information or restating it more explicitly if they believe the question is not clear. This step reveals how much ambiguity or underspecification is present in the input. The semantic changes between the paraphrased and clarified versions are used to estimate AU, reflecting how ambiguous the prompt may be.

\paragraph{Answering:} The model generates an initial answer based on the clarified question. This step reflects the model’s reasoning ability and knowledge without the benefit of reflection or revision.

\paragraph{Self-Checking:} In this stage, the model is asked to review and potentially revise its initial answer. OU is estimated by measuring the divergence between initial and self-checked answer distributions based on the idea that a knowledgeable model should be able to detect and correct generation level errors. EU is estimated using the entropy of the self-checked answer distribution. Persistent variability here indicates genuine knowledge gaps or reasoning limitations that remain even after self-verification.

By analyzing the outputs at each stage and comparing them across chains, we can estimate different uncertainty sources of the response. Detail prompt for each stage are shown in Appendix \ref{prompt_template}.

\subsection{Uncertainty Source Estimator}
\label{sec:source_est}
To disentangle the distinct uncertainty sources in LLM responses, we design four estimators, each aligned with a conceptual uncertainty source introduced in Section~\ref{sec:uncertainty_source_type}. These estimators operate over multiple response chains generated by our prompting framework, and are constructed to quantify uncertainty at specific stages of the question-answering process. Let $N$ denote the number of sampled chains.

\paragraph{Surface Form Uncertainty (SU)} This source reflects the model's sensitivity to superficial linguistic variation in input phrasing. We compute the average lexical distance between the original question $q_0$ and its paraphrased variants $\{q_p^{(i)}\}_{i=1}^N$ using a divergence metric $\text{LexDist}(\cdot,\cdot)$ (e.g., negative ROUGE or BLEU): $\text{SU}(q_0, \{q_p^{(i)}\}) = \frac{1}{N} \sum_{i=1}^N \text{LexDist}(q_0, q_p^{(i)})$.

\paragraph{Aleatoric Uncertainty (AU)} AU captures uncertainty from input ambiguity or underspecification. It is defined as the mean semantic distance between paraphrased questions and their clarifications: $\text{AU}(\{q_p^{(i)}, q_c^{(i)}\}) = \frac{1}{N} \sum_{i=1}^N \text{SemDist}(q_p^{(i)}, q_c^{(i)})$, where $\text{SemDist}$ is a semantic distance metric (e.g., cosine distance, L2 distance).

\paragraph{Epistemic Uncertainty (EU)} EU reflects knowledge limitations or reasoning failure, quantified by the entropy of self-checked answers: $\text{EU} = \mathcal{H}(P_{\text{SC}}) = - \sum_{a \in \mathcal{A}} P_{\text{SC}}(a) \log P_{\text{SC}}(a)$, where $P_{\text{SC}}$ is the empirical distribution over self-checked answers.

\paragraph{Operational Uncertainty (OU)} OU captures inconsistencies due to sampling stochasticity or execution errors. It is defined as the Jensen-Shannon divergence between the first-trial and self-checked answer distributions: $\text{OU} = D_{\text{JS}}(P_{\text{SC}} \,\|\, P_{\text{FT}})$.

\paragraph{Estimator Validity} We acknowledge that these estimators may not yield complete source disentanglement, as uncertainty can propagate between stages of the prompting pipeline or be confounded by interactions between sources. Nevertheless, we posit that the design of each estimator aligns tightly with the behavioral signature of its corresponding uncertainty source. This conceptual alignment, combined with empirical validation in downstream prediction tasks, supports the interpretive and practical utility of the proposed estimators.

\subsection{Uncertainty Source Estimator Validation through Predictive Performance}
\label{sec:source_est_validation}

To evaluate the validity of our uncertainty source estimators, we measure their predictive performance across three benchmark datasets—\textsc{CommonsenseQA}~\citet{talmor2018commonsenseqa}, \textsc{GSM8K}~\citet{cobbe2021gsm8k}, and \textsc{TruthfulQA}~\citet{lin2021truthfulqa}—using four language models from two families~\citet{grattafiori2024llama,team2024gemma}: \textsc{LLaMA-3.2-3B}, \textsc{LLaMA-3-8B}, \textsc{Gemma-2-2B}, and \textsc{Gemma-2-9B}. Specifically, we assess how well each estimator can distinguish between correct and incorrect model responses, using metrics such as AUROC and AUPRC~\citet{qi2021stochastic}. Strong predictive performance indicates that the estimator captures uncertainty signals that are meaningfully correlated with actual model errors, supporting its reliability for downstream applications. In contrast, poor predictive performance suggests misalignment between the estimator and the type of uncertainty it aims to represent, limiting its usefulness for interpretability or decision-making.

We frame the prediction of each uncertainty source as a binary classification task, where the goal is to identify whether a given question exhibits significant uncertainty of a specific type. Specifically, we generate 32 responses for each question and compute its empirical accuracy. Questions with accuracy lower than 70\% are labeled as \textit{uncertain} (positive class). This setup allows us to evaluate how well each uncertainty source estimator predicts low-confidence or error-prone questions that are plausibly driven by the respective uncertainty factor. We evaluate each estimator using two threshold-independent metrics: Area Under the Receiver Operating Characteristic curve (AUROC) and Area Under the Precision-Recall Curve, both averaged across all four models. Full per-model performance and class imbalance ratios are provided in Appendix~\ref{source_est_performance_detail}.

As shown in Table~\ref{tab:source_est_performance}, the EU estimator consistently achieves strong AUROC and AUPRC scores, highlighting its effectiveness in capturing knowledge-related uncertainty aligned with correctness. OU also performs well, especially on \textsc{GSM8K}, where multi-step reasoning introduces execution errors. In contrast, SU and AU exhibit lower performance in \textsc{CommonsenseQA} and \textsc{TruthfulQA}, likely because the questions are short and unambiguous, limiting both surface-level variation and the potential for ambiguity. Still, they remain relevant for real-world use cases involving short or ambiguous inputs. Overall, each estimator offers complementary insight into different uncertainty sources and LLM failure modes.

\begin{table}[htbp]
  \caption{\textbf{Predictive Performance of Uncertainty Source Estimators.} 
AUROC and AUPRC scores (averaged over four models) indicate how well each estimator identifies uncertain questions across three benchmarks. EU and OU estimators perform consistently well, while SU and AU estimators show weaker signals likely due to the clarity and brevity of the dataset.}

  \label{tab:source_est_performance}
  \centering
  \begin{tabular}{lcccccc}
    \toprule
    \multirow{2}{*}{Estimator} &
    \multicolumn{2}{c}{\textbf{CommonsenseQA}} &
    \multicolumn{2}{c}{\textbf{GSM8K}} &
    \multicolumn{2}{c}{\textbf{TruthfulQA}} \\
    \cmidrule(r){2-3} \cmidrule(r){4-5} \cmidrule(r){6-7}
    & AUROC & AUPRC & AUROC & AUPRC & AUROC & AUPRC \\
    \midrule
    SU Estimator & 0.514 & 0.389 & 0.611 & 0.455 & 0.504 & 0.379 \\
    AU Estimator & 0.526 & 0.377 & 0.471 & 0.386 & 0.448 & 0.360 \\
    EU Estimator & 0.868 & 0.781 & 0.949 & 0.915 & 0.955 & 0.933 \\
    OU Estimator  & 0.552 & 0.489 & 0.837 & 0.733 & 0.717 & 0.611 \\
    \bottomrule
  \end{tabular}%
\end{table}

\section{Profiling Uncertainty Across Tasks, Models, and Metrics}
\label{sec:profile}
In this section, we investigate how different uncertainty sources are distributed across tasks, models, and existing LLM uncertainty metrics from previous works. Understanding these distributions helps us profile how uncertainty manifests in different settings, laying the foundation for applications such as task-specific metric and model selection shown in section \ref{sec:application}.

\subsection{Implementation}
\label{sec:profile:implementation}
\paragraph{Datasets and Models} We profile uncertainty using multiple datasets and models spanning diverse task types and model families. Full usage details are provided in Appendix~\ref{app:profile_setup}.

\paragraph{Profiling Existing Uncertainty Metrics} We benchmark eight task-agnostic uncertainty metrics that don't require additional model training. These metrics can be categorized into four types. Token-liklihood based metrics estimate uncertainty from the predictive entropy of model’s output distribution, including \textsc{Normalized Predictive Entropy (NPE)}, \textsc{Length-Normalized Predictive Entropy (LNPE)} \citet{malinin2020uncertainty}, and \textsc{Semantic Entropy (SE)} \citet{kuhn2023semantic}. Verbalized-based measures assess confidence by directly prompting the model to express its certainty, such as \textsc{Verbalized Confidence (VC)} and \textsc{P(True)} \citet{kadavath2022language}. Lexical-based metrics, such as \textsc{Lexical Similarity }\citet{lin2022towards}, evaluate uncertainty based on the consistency of lexical overlap between multiple responses. EU-AU specific metrics target epistemic and aleatoric uncertainty explicitly through input perturbation or conditional iterative prompting, including \textsc{SPUQ} \citet{gao2024spuq} and \textsc{IPT-EU} \citet{yadkori2024believe}. 
Since \textsc{SPUQ}, \textsc{P(True)}, and \textsc{VC} are originally formulated to express model confidence, we transform them into uncertainty measures by taking their complements: specifically, \textsc{SPUQ-Comp}, \textsc{PTrue-Comp}, and \textsc{VC-Neg} represent the complement of SPUQ, the complement of P(True), and the negative of VC, respectively.

\begin{figure}[t]
\vspace{-1.3cm}
    \centering
    \makebox[\textwidth]{%
    \resizebox{1.0\linewidth}{!}{%
        \begin{tabular}{c}
            \includegraphics{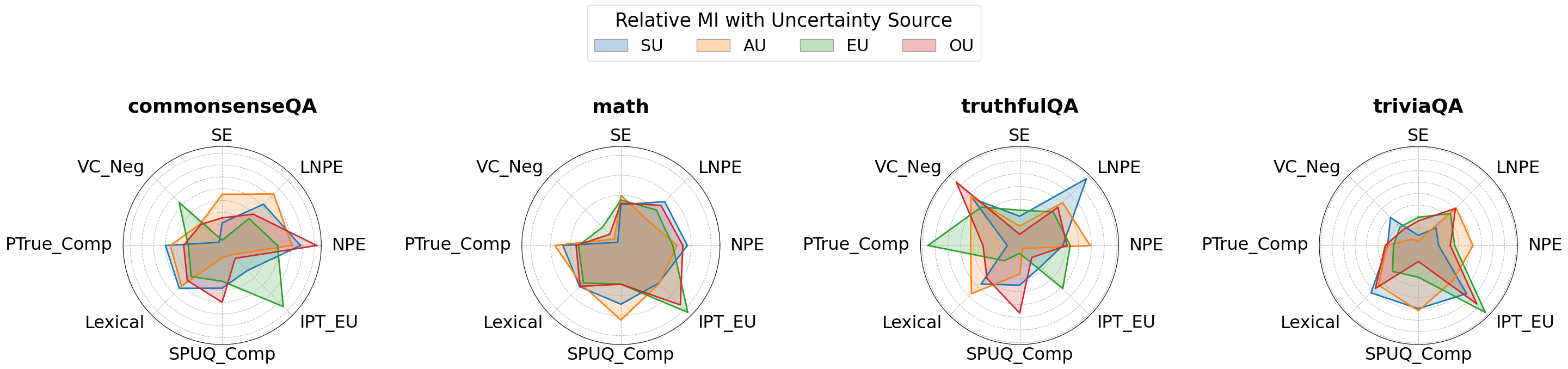}
        \end{tabular}
    }
    }
    \caption{\textbf{Uncertainty Characteristic of Existing Metrics.} The figure shows the relative magnitude of mutual information between each uncertainty source and the evaluated metrics, normalized within each source. A higher peak for a given uncertainty source indicates that the corresponding metric is more strongly influenced by that source compared to others. }
    \label{fig:est_source_mi}
\end{figure}

\paragraph{Analytical Approach} To profile uncertainty characteristics across different dimensions, we adopt three complementary analyses.

First, we compute the mutual information between each uncertainty source estimator and a set of evaluation metrics to assess how well each metric captures specific types of uncertainty. Mutual information is a general-purpose statistical measure that quantifies the dependency between two random variables, without assuming a linear relationship or specific distributional form. This makes it particularly well-suited for analyzing non-monotonic or complex associations between uncertainty estimates and metric scores.
Formally, the mutual information between random variables $U$ (uncertainty estimate) and $M$ (metric score) is defined as:
$
I(U; M) = \sum_{u \in \mathcal{U}} \sum_{m \in \mathcal{M}} P(u, m) \log \left( \frac{P(u, m)}{P(u) P(m)} \right),
$
where $P(u, m)$ is the joint probability distribution, and $P(u)$ and $P(m)$ are the marginal distributions. Higher mutual information values indicate stronger statistical dependence, suggesting that the metric is more sensitive to variation in that specific source of uncertainty. 

Second, to examine how uncertainty varies by task, we compute the average uncertainty source estimates across all models for each dataset. This reveals which types of uncertainty are most prominent in different task settings. Third, to assess model-level tendencies, we average uncertainty source estimates across all datasets for each model. Higher estimates indicate that the model is more susceptible to the corresponding source of uncertainty.

Details of the sampling protocol and statistical evaluation are provided in Appendix~\ref{sec:profile:stats}, and compute resources are described in Appendix~\ref{sec:profile:compute}.


\subsection{Discussion and Insights from Uncertainty Profiling}
In this section, we explore and discuss several key questions to understand how different types of uncertainty manifest themselves across metrics, tasks, and models. Additional analysis results are in Appendix \ref{additional_result}.

\paragraph{What type of uncertainty does each metric generally capture?} 
Figure~\ref{fig:est_source_mi} presents the mutual information between each benchmarked metric and the uncertainty source estimators across datasets. A dataset-wise breakdown reveals that uncertainty metrics behave differently depending on the task. For example, EU-specific metrics such as IPT-EU show significantly higher mutual information with the epistemic uncertainty estimator on datasets like \textsc{Math} and \textsc{TriviaQA}, where reasoning and factual knowledge play a central role. In contrast, metrics like VC-Neg consistently exhibit low mutual information across most uncertainty sources on \textsc{Math}, indicating limited capacity to capture complex reasoning-related uncertainties.

\begin{figure}[t]
\vspace{-1.3cm}
    \centering
    \begin{subfigure}[t]{0.48\linewidth}
        \centering
        \includegraphics[width=\linewidth]{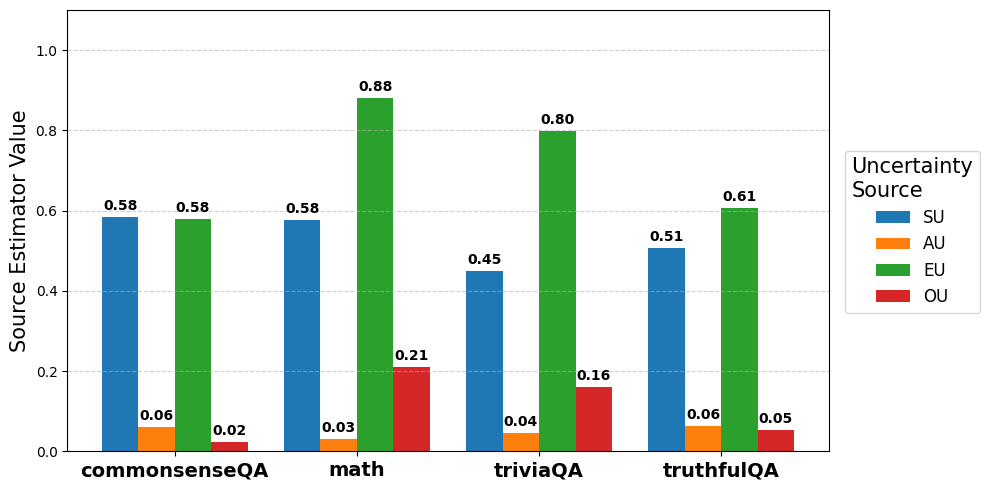}
        \caption{Average uncertainty source per dataset.}
        \label{fig:uncertainty-per-dataset}
    \end{subfigure}
    \hfill
    \begin{subfigure}[t]{0.48\linewidth}
        \centering
        \includegraphics[width=\linewidth]{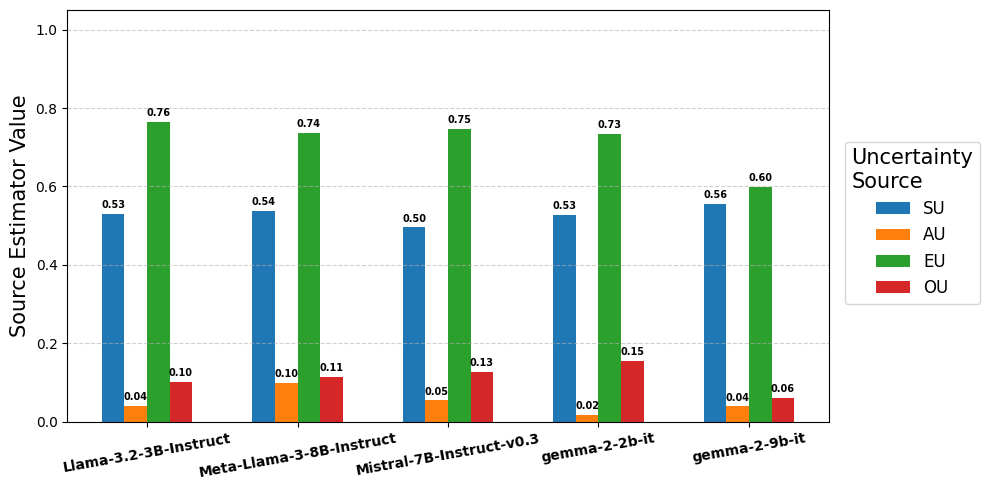}
        \caption{Average uncertainty source per model.}
        \label{fig:uncertainty-per-model}
    \end{subfigure}
    \caption{\textbf{Uncertainty Characteristics of Datasets and Models.} 
    (Left) SU and AU are relatively consistent across datasets, while EU and OU vary more significantly, particularly elevated in \textsc{Math} and \textsc{TriviaQA}. 
    (Right) Uncertainty profiles remain fairly stable across model sizes, with SU and EU being the dominant contributors, while AU and OU are generally less pronounced.}
    \label{fig:uncertainty-analysis}
    \vspace{-0.1cm}
\end{figure}

\paragraph{Do certain tasks consistently exhibit stronger tendencies toward specific uncertainty source?}
Figure~\ref{fig:uncertainty-per-dataset} shows the average values of each uncertainty source estimate across datasets. Overall, SU and EU emerge as the most prominent sources across tasks. In contrast, OU is particularly pronounced in \textsc{Math} and \textsc{TriviaQA}, while AU remains consistently low across all datasets. 

These patterns are consistent with prior research, which has identified similar challenges related to uncertainty in language model performance. High SU scores suggest that models are sensitive to input phrasing, implying that techniques like prompt optimization \citet{khattab2024dspy, yang2023large} or paraphrasing \citet{zhou2024paraphrase} could significantly impact performance. EU tends to be higher in tasks requiring complex reasoning(\textsc{Math}) or reading comprehension (\textsc{TriviaQA}). These tasks often involve open-ended or essay-style questions rather than multiple-choice formats (\textsc{CommonsenseQA}, \textsc{TruthfulQA}), which places greater demands on the model to generate accurate and complete responses from scratch \citet{yadkori2024believe}. OU is also elevated in these datasets, likely because longer, multi-step generative answers are more prone to execution or formatting errors compared to shorter, constrained outputs \citet{xu2025cod, zeng2025revisiting}. In contrast, AU is relatively low across all datasets, likely because the benchmark questions are carefully curated to reduce ambiguity.

\paragraph{Are different language models more susceptible to specific uncertainty source?}
Figure~\ref{fig:uncertainty-per-model} presents the average values of each uncertainty source estimate across models. Overall, the distribution of uncertainty sources is relatively consistent across models, with SU and EU being the dominant contributors, while AU and OU remain lower. This suggests that, regardless of model size, surface-level sensitivity and knowledge-related uncertainty are the most prominent challenges.

Interestingly, there is no clear distinction between smaller and larger models in terms of their average uncertainty source values. This contrasts with the intuitive expectation that smaller models would exhibit higher uncertainty, leading to lower accuracy. One possible explanation is that, while the estimated uncertainty levels may be similar, the tolerance for uncertainty differs by model size. A small amount of uncertainty value change might significantly degrade the performance of smaller models, whereas larger models may be more robust to the same level of uncertainty due to stronger internal representations or reasoning capabilities.

\begin{figure}[h]
    \centering
    \includegraphics[width=1.0\linewidth]{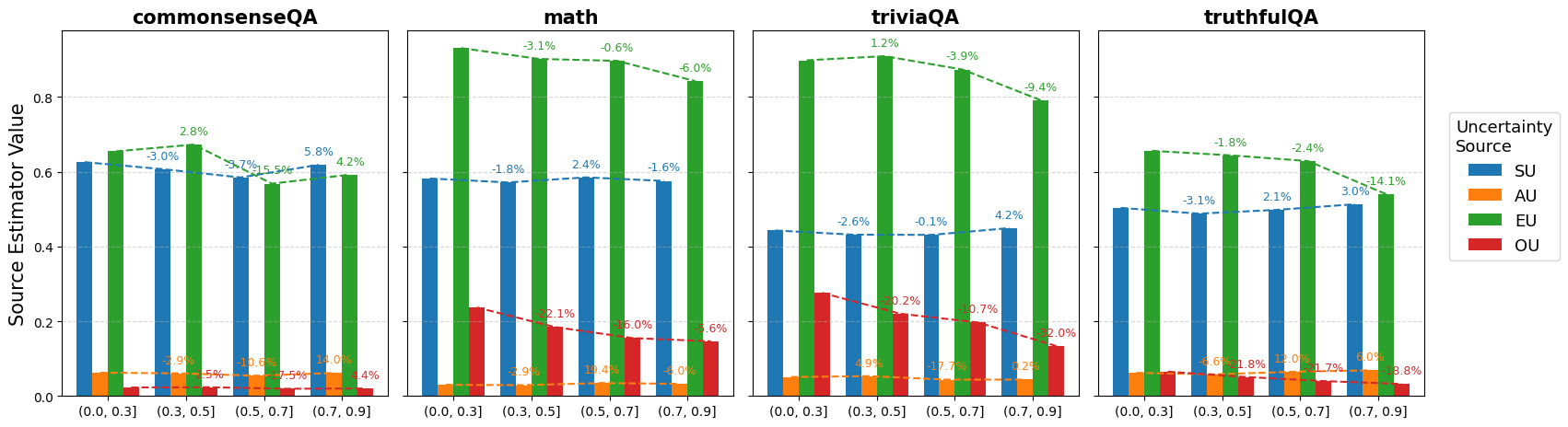}
    \caption{\textbf{Uncertainty Source Variation by Question Difficulty.} Each plot shows average uncertainty values across difficulty bins. EU and OU decrease with easier questions, indicating greater reducibility. SU and AU remain stable, suggesting more intrinsic or input-dependent variability.}
    \label{fig:uncertainty-difficulty}
    \vspace{-0.1cm}
\end{figure}

\paragraph{How does task difficulty relate to uncertainty source distribution?}
Figure~\ref{fig:uncertainty-difficulty} illustrates how the values of different uncertainty sources vary with question difficulty. We divide each dataset into four bins based on average response accuracy to approximate difficulty levels. In \textsc{Math}, \textsc{TruthfulQA}, and \textsc{TriviaQA}, we observe a consistent decline in EU and OU as questions become easier. This suggests that these sources are more sensitive to task difficulty and potentially more reducible with improved model performance. In contrast, SU and AU remain relatively stable across difficulty levels, indicating they may reflect more intrinsic or input-dependent variability that is harder to reduce.

\section{Uncertainty Profile-Guided Adaptive Metric and Model Selection}
\label{sec:application}
\begin{figure}[t]
\vspace{-1.3cm}
    \centering
    \includegraphics[width=1.0\linewidth]{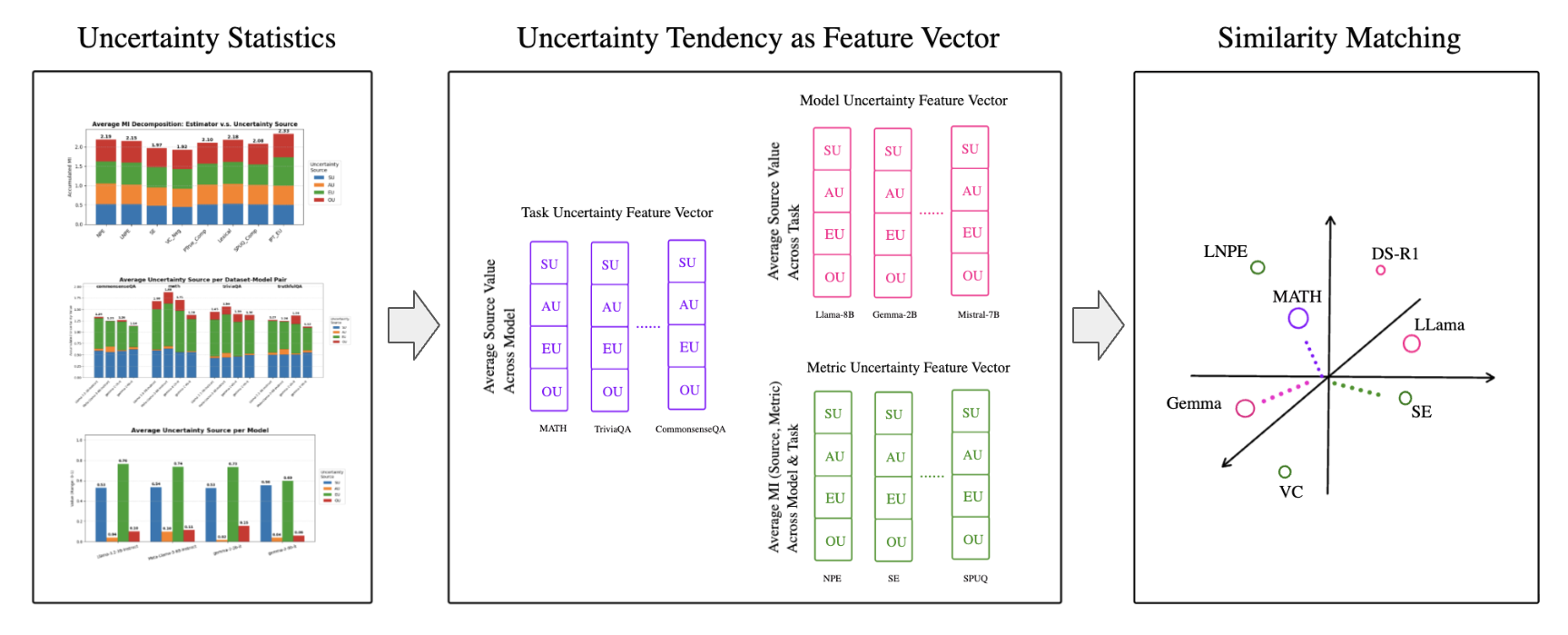}
    \caption{\textbf{Adaptive Metric and Model Selection Based on Uncertainty Profiles.} Figure illustrates the steps involved in selecting task-specific metrics/models: (1) Convert statistical results into uncertainty profile vectors; (2) Compute the similarity between the task profile and the profiles of potential candidate metrics/models; (3) Select the most compatible metric/model based on the dominant uncertainty characteristics of the task.}

    \label{fig:apply}
\end{figure}

In this section, we present a framework that uses uncertainty source profiles to adaptively select evaluation metrics and models based on task characteristics. As shown in Figure~\ref{fig:apply}, we convert results from Section~\ref{sec:profile} into uncertainty profile vectors, then compute their similarity with candidate metrics or models. This enables task-aware selection aligned with dominant uncertainty patterns. Details of vector conversion and normalization are provided in Appendix~\ref{sec:profile:convert}.

\subsection{Experiment Setup}
\label{sec:application_exp}
We evaluate our framework across three scenarios, demonstrating consistent gains over non-adaptive baselines.

\paragraph{Scenario 1: \textbf{Metric Selection for a Task}}  
We identify the most suitable evaluation metric for a given task by comparing uncertainty profiles. Each dataset is represented as a \textsc{Dataset-Vec} (Figure~\ref{fig:uncertainty-per-dataset}) and each metric as a \textsc{Metric-Vec} (Figure~\ref{fig:est_source_mi}), based on mutual information with uncertainty sources. Metrics with the highest cosine similarity to the task vector are selected.

\paragraph{Scenario 2: \textbf{Model Selection for a Task}}  
Here, we select the model likely to perform best on a given dataset. Each model has a \textsc{Model-Vec} summarizing its average uncertainty behavior. We hypothesize that models whose uncertainty profiles are less similar to the task’s profile (\textsc{Dataset-Vec}) are more robust to those uncertainty sources and therefore better suited for the task.

\paragraph{Scenario 3: \textbf{Metric Selection for Task and Model}}  
We jointly consider both task and model profiles to select the most aligned metric. Each metric's similarity to the task and model vectors is computed, and their geometric mean is used to prioritize metrics that align well with both, penalizing those aligned with only one.

\paragraph{Evaluation} We evaluate our method using the same four datasets and five models as in Section~\ref{sec:profile:implementation}. Each scenario is treated as a ranking task, aiming to prioritize the most appropriate metric or model. We use Normalized Discounted Cumulative Gain (NDCG) as the evaluation metric, computing NDCG(@all) over the full ranked list. Accuracy is used as the relevance score for model selection, and AUROC is used for metric selection. NDCG is ideal here as it emphasizes correctly ranking highly relevant items, accounts for non-binary relevance, and penalizes large ranking errors more strongly than minor ones.

\subsection{Experiment Result and Analysis}
\label{sec:application_ra}

To test our method, we include both the worst-case scenario and the average NDCG score from 100 random samplings as baselines. As shown in Table~\ref{tab:ndcg-combined}, our uncertainty-guided selection strategy consistently outperforms both the worst‑case and the random baselines across all three evaluation scenarios. While in some cases, random selection already produces a high NDCG scores of above 90\%, our method still achieves average gains of 3.7\% in Scenario 1 and Scenario 3, demonstrating its robustness. In Scenario 2, our strategy achieves a substantial average gain of 5.5\%.


\begin{table}[htbp]
  \caption{\textbf{NDCG score (NDCG@all) across three evaluation scenarios.} \textsc{Gain} indicates the percentage improvement over the random selection baseline. }
  \label{tab:ndcg-combined}
  \centering
  \resizebox{\textwidth}{!}{
    \begin{tabular}{lcccc|cccc|cccc}
      \toprule
      \multirow{2}{*}{\textbf{Dataset}} & \multicolumn{4}{c|}{\textbf{Scenario 1}} & \multicolumn{4}{c|}{\textbf{Scenario 2}} & \multicolumn{4}{c}{\textbf{Scenario 3}} \\
      \cmidrule(lr){2-5} \cmidrule(lr){6-9} \cmidrule(lr){10-13}
      & Ours & Worst & Random & Gain & Ours & Worst & Random & Gain & Ours & Worst & Random & Gain \\
      \midrule
      CommonsenseQA & \textbf{0.973} & 0.840 & 0.933 & +4.1\% & \textbf{0.878} & 0.673 & 0.846 & +3.8\% & \textbf{0.967} & 0.837 & 0.930 & +3.9\% \\
      Math          & \textbf{0.954} & 0.795 & 0.924 & +3.3\% & \textbf{0.881} & 0.640 & 0.825 & +6.8\% & \textbf{0.951} & 0.789 & 0.919 & +3.5\% \\
      TriviaQA      & \textbf{0.968} & 0.826 & 0.932 & +3.8\% & \textbf{0.986} & 0.919 & 0.955 & +3.2\% & \textbf{0.965} & 0.822 & 0.929 & +3.9\% \\
      TruthfulQA    & \textbf{0.969} & 0.863 & 0.936 & +3.6\% & \textbf{0.811} & 0.564 & 0.751 & +8.0\% & \textbf{0.965} & 0.856 & 0.933 & +3.4\% \\
      \midrule
      Average       & \textbf{0.966} & 0.831 & 0.931 & +3.7\% & \textbf{0.899} & 0.699 & 0.844 & +5.5\% & \textbf{0.962} & 0.826 & 0.928 & +3.7\% \\
      \bottomrule
    \end{tabular}
  }
\end{table}

\begin{figure}[htbp]
    \centering
    \begin{subfigure}[t]{0.48\linewidth}
        \centering
        \includegraphics[width=\linewidth]{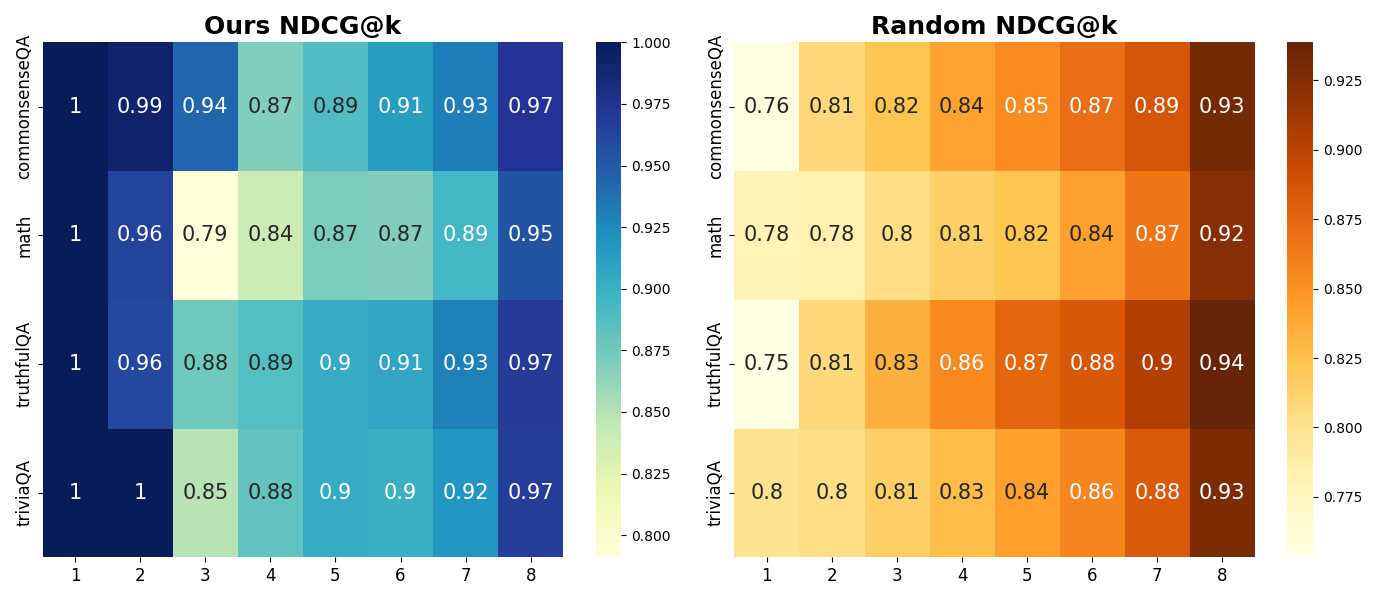}
        \caption{NDCG@K for Scenario 1}
        \label{fig:s1-ndcg@k}
    \end{subfigure}
    \hfill
    \begin{subfigure}[t]{0.48\linewidth}
        \centering
        \includegraphics[width=\linewidth]{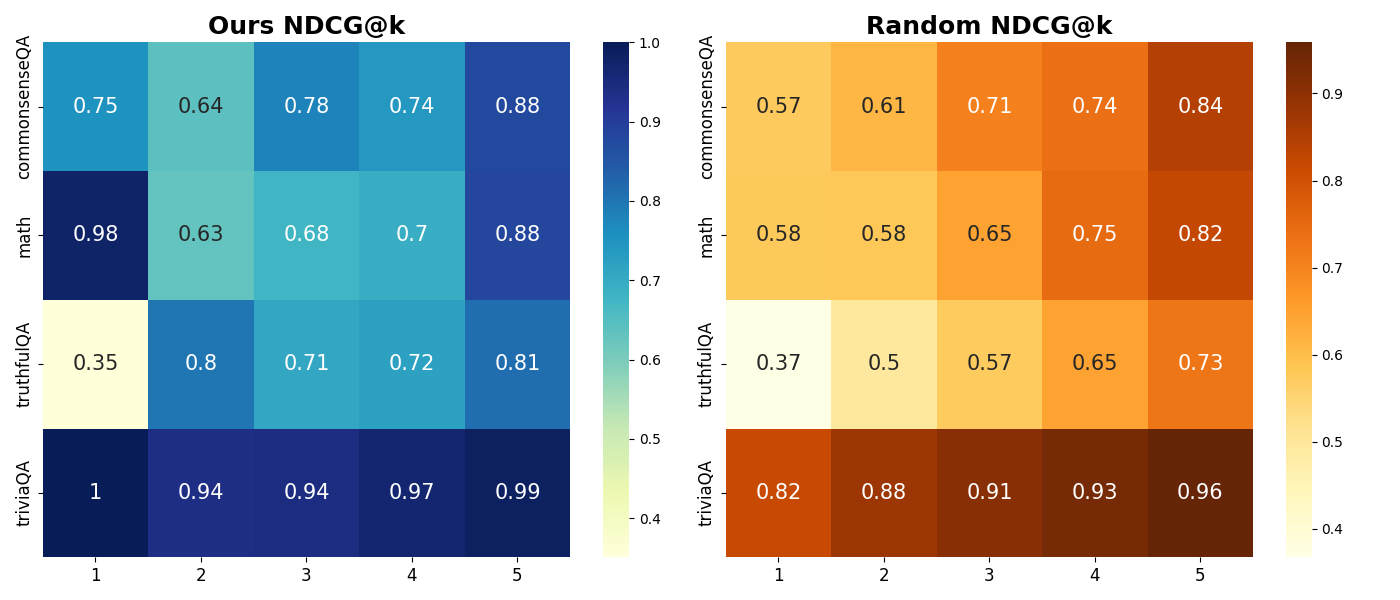}
        \caption{NDCG@K for Scenario 2}
        \label{fig:s2-ndcg@k}
    \end{subfigure}
    \caption{\textbf{Performance at different cutoff ranks.} Heatmaps show \textsc{NDCG@K} values for (a) Scenario 1 and (b) Scenario 2 of our method and random selection. Our method performs well at lower cutoff ranks (e.g., \(K=1\) or \(K=2\)), but experiences a gradual decline at higher ranks, reflecting a diminishing ability to maintain relevance as more items are considered.}
    \label{fig:ndcg@k}
\end{figure}

Heatmaps in Figure~\ref{fig:ndcg@k} show the NDCG scores at different cutoff ranks \(K\) for our method and a random baseline in Scenario 1 and 2. Our method performs strongly at lower \(K\), indicating that it can rank the most relevant items at the top. However, performance gradually declines at higher \(K\), suggesting that its ability to maintain relevance decreases as more items are considered. This decline is likely due to two factors: First, our method matches models and metrics based on uncertainty characteristics, but mismatches between uncertainty estimates and actual correctness can still occur, as discussed in Section~\ref{sec:profile}. Second, at higher K, the long tail of less distinguishable items diminishes the impact of uncertainty, reducing our method's advantage over random selection. 

While our method doesn't perfectly optimize for all ranks, it consistently provides meaningful improvements without requiring additional training or supervision, making it a practical and generalizable alternative to manual tuning or trial-and-error approach in real-world applications. Future work will explore deeper on the mappings between uncertainty characteristic and performance, with the goal of enhancing ranking stability and extending applicability to more complex scenarios.

\section{Conclusion}
\label{sec:conclusion}

We present a framework that decomposes LLM uncertainty into four interpretable sources and quantifies each using dedicated estimators. By profiling these sources across tasks, models, and metrics, we uncover meaningful patterns that inform uncertainty-aware metric and model selection. Our adaptive selection method, driven by uncertainty profiles, consistently outperforms non-adaptive baselines. This work offers a step toward more interpretable and task-aware evaluation of LLM behavior and offers insights for improving LLM performance in various practical applications.

\bibliography{neurips_2025}           

\appendix

\section{Implementation Detail}
\label{app:implementation-detail}

\subsection{Uncertainty Source Estimator}
\label{sec:source_est_implement}
\paragraph{AU \& SU} In our following experiments. For AU, we use the hidden state of the model last layer as semantic embedding and use cosine distance as the semantic distance metric. For SU, we choose Rouge-l as the lexical distance measure. We also conduct experiments on the selection of other distance metric and the results are in appendix \ref{su_au_est_selection}.

\paragraph{Normalization and Scaling} To ensure that all four uncertainty estimators are comparable and interpretable within a unified $[0,1]$ range, we apply appropriate scaling functions to each raw metric. SU, based on lexical distance, is naturally bounded between 0 and 1 and requires no additional scaling. AU, measured as cosine distance, is divided by 2 to ensure its values fall within the desired range. EU, derived from entropy, is scaled using the transformation $1-2^{-\text{EU}}$ to compress its potentially unbounded values while preserving interpretability. OU, computed as Jensen-Shannon divergence, is normalized by dividing by $\log 2$, which is its theoretical maximum. This normalization facilitates cross-source comparison and improves the stability of downstream evaluation and visualization.

\subsection{Uncertainty Profile Vector Conversion}
\label{sec:profile:convert}
For each scenario, since we rank a set of model or metric vectors based on a given vector (or vectors), the relative differences between them matter more than their absolute values. Therefore, we apply column-wise min-max scaling to convert the original uncertainty source values into relative values before computing cosine similarity. This normalization allows us to focus purely on the relative alignment between vectors.

\section{Experiment Detail}

\subsection{Datasets and Models}
\label{app:profile_setup}

To capture uncertainty profiles across tasks, models, and metrics, we select four diverse datasets: \textsc{MATH} (mathematical reasoning)~\citet{hendrycksmath2021}, \textsc{CommonsenseQA} (commonsense reasoning)~\citet{talmor2018commonsenseqa}, \textsc{TriviaQA} (reading comprehension)~\citet{2017arXivtriviaqa}, and \textsc{TruthfulQA} (safety and truthfulness)~\citet{lin2021truthfulqa}. These datasets are evaluated using five open-source language models of varying sizes: \textsc{LLaMA-3.2-3B}, \textsc{LLaMA-3-8B}, \textsc{Gemma-2-2B}, \textsc{Gemma-2-9B}, and \textsc{Mistral-7B-v0.3}. This selection enables broad coverage of task types and architectural characteristics.

\subsection{Compute Resources}
\label{sec:profile:compute}
All experiments are conducted using a single NVIDIA GeForce RTX 3090 GPU. We use the vLLM engine for efficient inference and extract hidden states from quantized models hosted on Hugging-Face Hub.

\subsection{Statistical Significance and Evaluation Protocol}
\label{sec:profile:stats}
Due to computational constraints, we randomly sample 150 questions from each dataset. To ensure robustness and statistical significance, we apply bootstrapping with 500 resamples across all experiments. For benchmark metric evaluations, we use: 32 samples for \textsc{NPE}, \textsc{LNPE}, and \textsc{SE}; 3 samples for \textsc{VC-Neg} and \textsc{PTrue-Comp}; 5 paraphrased variants for \textsc{SPUQ-Comp}; and 4 iterative chains (depth 5) for \textsc{IPT-EU}. Questions with accuracy below 70\% are labeled as uncertain (positive label).

\section{Source Estimator Benchmarking Result}
\label{source_est_performance_detail}

\begin{table}[H]
  \caption{AU Estimator Performance}
  \label{tab:au_est_performance}
  \centering
  \resizebox{\textwidth}{!}{%
  \begin{tabular}{lcccccccccccccc}
    \toprule
    \multirow{2}{*}{Dataset} &
    \multicolumn{3}{c}{Llama-3.2-3B} &
    \multicolumn{3}{c}{Llama-3-8B} &
    \multicolumn{3}{c}{Gemma-2-2B} &
    \multicolumn{3}{c}{Gemma-2-9B} \\
    \cmidrule(r){2-4} \cmidrule(r){5-7} \cmidrule(r){8-10} \cmidrule(r){11-13}
    & AUROC & AUPRC & P/N & AUROC & AUPRC & P/N & AUROC & AUPRC & P/N & AUROC & AUPRC & P/N \\
    \midrule
    commonsenseQA & 0.472 & 0.346 & 44:56 & 0.537 & 0.334 & 76:24 & 0.516 & 0.344 & 63:37 & 0.580 & 0.484 & 27:73 \\
    gsm8k         & 0.443 & 0.371 & 30:70 & 0.453 & 0.328 & 44:56 & 0.433 & 0.306 & 57:43 & 0.554 & 0.539 & 23:77 \\
    truthfulQA    & 0.421 & 0.276 & 59:41 & 0.514 & 0.354 & 48:52 & 0.404 & 0.297 & 48:52 & 0.452 & 0.513 & 28:72 \\
    \bottomrule
  \end{tabular}%
  }
\end{table}

\begin{table}[H]
  \caption{SU Estimator Performance}
  \label{tab:su_est_performance}
  \centering
  \resizebox{\textwidth}{!}{%
  \begin{tabular}{lcccccccccccccc}
    \toprule
    \multirow{2}{*}{Dataset} &
    \multicolumn{3}{c}{Llama-3.2-3B} &
    \multicolumn{3}{c}{Llama-3-8B} &
    \multicolumn{3}{c}{Gemma-2-2B} &
    \multicolumn{3}{c}{Gemma-2-9B} \\
    \cmidrule(r){2-4} \cmidrule(r){5-7} \cmidrule(r){8-10} \cmidrule(r){11-13}
    & AUROC & AUPRC & P/N & AUROC & AUPRC & P/N & AUROC & AUPRC & P/N & AUROC & AUPRC & P/N \\
    \midrule
    commonsenseQA & 0.559 & 0.433 & 42:58 & 0.509 & 0.314 & 95:5 & 0.520 & 0.393 & 75:25 & 0.466 & 0.414 & 26:74 \\
    gsm8k         & 0.649 & 0.467 & 33:67 & 0.579 & 0.433 & 44:56 & 0.598 & 0.395 & 58:42 & 0.617 & 0.526 & 22:78 \\
    truthfulQA    & 0.499 & 0.305 & 58:42 & 0.538 & 0.377 & 45:55 & 0.491 & 0.306 & 43:57 & 0.486 & 0.526 & 25:75 \\
    \bottomrule
  \end{tabular}%
  }
\end{table}

\begin{table}[H]
  \caption{EU Estimator Performance}
  \label{tab:eu_est_performance}
  \centering
  \resizebox{\textwidth}{!}{%
  \begin{tabular}{lcccccccccccccc}
    \toprule
    \multirow{2}{*}{Dataset} &
    \multicolumn{3}{c}{Llama-3.2-3B} &
    \multicolumn{3}{c}{Llama-3-8B} &
    \multicolumn{3}{c}{Gemma-2-2B} &
    \multicolumn{3}{c}{Gemma-2-9B} \\
    \cmidrule(r){2-4} \cmidrule(r){5-7} \cmidrule(r){8-10} \cmidrule(r){11-13}
    & AUROC & AUPRC & P/N & AUROC & AUPRC & P/N & AUROC & AUPRC & P/N & AUROC & AUPRC & P/N \\
    \midrule
    commonsenseQA & 0.896 & 0.769 & 39:61 & 0.890 & 0.841 & 75:25 & 0.811 & 0.679 & 67:33 & 0.873 & 0.835 & 25:75 \\
    gsm8k         & 0.939 & 0.878 & 31:69 & 0.963 & 0.914 & 42:58 & 0.969 & 0.942 & 61:39 & 0.925 & 0.927 & 23:77 \\
    truthfulQA    & 0.965 & 0.927 & 58:42 & 0.956 & 0.919 & 43:57 & 0.973 & 0.943 & 44:56 & 0.927 & 0.944 & 27:73 \\
    \bottomrule
  \end{tabular}%
  }
\end{table}

\begin{table}[H]
  \caption{OU Estimator Performance}
  \label{tab:ou_est_performance}
  \centering
  \resizebox{\textwidth}{!}{%
  \begin{tabular}{lcccccccccccccc}
    \toprule
    \multirow{2}{*}{Dataset} &
    \multicolumn{3}{c}{Llama-3.2-3B} &
    \multicolumn{3}{c}{Llama-3-8B} &
    \multicolumn{3}{c}{Gemma-2-2B} &
    \multicolumn{3}{c}{Gemma-2-9B} \\
    \cmidrule(r){2-4} \cmidrule(r){5-7} \cmidrule(r){8-10} \cmidrule(r){11-13}
    & AUROC & AUPRC & P/N & AUROC & AUPRC & P/N & AUROC & AUPRC & P/N & AUROC & AUPRC & P/N \\
    \midrule
    commonsenseQA & 0.355 & 0.265 & 39:61 & 0.523 & 0.406 & 75:25 & 0.645 & 0.562 & 67:33 & 0.684 & 0.721 & 25:75 \\
    gsm8k         & 0.675 & 0.511 & 30:70 & 0.889 & 0.759 & 42:58 & 0.907 & 0.767 & 61:39 & 0.875 & 0.895 & 23:77 \\
    truthfulQA    & 0.696 & 0.508 & 58:42 & 0.694 & 0.513 & 43:57 & 0.729 & 0.582 & 44:56 & 0.747 & 0.840 & 27:73 \\
    \bottomrule
  \end{tabular}%
  }
\end{table}

\section{SU AU Distance Metric Selection}
\label{su_au_est_selection}

\begin{table}[H]
  \centering
  \caption{AU Estimator Semantic Distance Performance Comparison.}
  \label{tab:au_distance_metrics}
  \resizebox{\textwidth}{!}{%
  \begin{tabular}{llcccccc}
    \toprule
    Dataset & Model & \multicolumn{2}{c}{Cosine} & \multicolumn{2}{c}{L1} & \multicolumn{2}{c}{L2} \\
    \cmidrule(r){3-4} \cmidrule(r){5-6} \cmidrule(r){7-8}
    & & AUROC & AUPR & AUROC & AUPR & AUROC & AUPR \\
    \midrule
    commonsenseQA & Llama-3.2-3B-Instruct & 0.472 & 0.346 & 0.465 & 0.372 & 0.471 & 0.383 \\
                  & Meta-Llama-3-8B-Instruct & 0.537 & 0.334 & 0.543 & 0.336 & 0.530 & 0.330 \\
                  & gemma-2-2b-it & 0.516 & 0.344 & 0.485 & 0.337 & 0.502 & 0.349 \\
                  & gemma-2-9b-it & 0.580 & 0.484 & 0.608 & 0.509 & 0.580 & 0.489 \\
    \midrule
    truthfulQA & Llama-3.2-3B-Instruct & 0.421 & 0.276 & 0.426 & 0.271 & 0.422 & 0.271 \\
               & Meta-Llama-3-8B-Instruct & 0.514 & 0.354 & 0.482 & 0.329 & 0.496 & 0.340 \\
               & gemma-2-2b-it & 0.404 & 0.297 & 0.417 & 0.293 & 0.419 & 0.293 \\
               & gemma-2-9b-it & 0.452 & 0.513 & 0.449 & 0.515 & 0.430 & 0.501 \\
    \midrule
    gsm8k & Llama-3.2-3B-Instruct & 0.443 & 0.371 & 0.457 & 0.385 & 0.458 & 0.386 \\
          & Meta-Llama-3-8B-Instruct & 0.453 & 0.328 & 0.444 & 0.345 & 0.444 & 0.347 \\
          & gemma-2-2b-it & 0.433 & 0.306 & 0.405 & 0.280 & 0.426 & 0.288 \\
          & gemma-2-9b-it & 0.554 & 0.539 & 0.526 & 0.528 & 0.534 & 0.531 \\
    \bottomrule
  \end{tabular}
  }
\end{table}

\begin{table}[H]
  \centering
  \caption{SU Estimator Lexical Distance Performance Comparison.}
  \label{tab:su_distance_metrics}
  \resizebox{\textwidth}{!}{%
  \begin{tabular}{llcccccc}
    \toprule
    Dataset & Model & \multicolumn{2}{c}{ROUGE-5} & \multicolumn{2}{c}{BLEU-5} & \multicolumn{2}{c}{ROUGE-L} \\
    \cmidrule(r){3-4} \cmidrule(r){5-6} \cmidrule(r){7-8}
    & & AUROC & AUPR & AUROC & AUPR & AUROC & AUPR \\
    \midrule
    commonsenseQA & Llama-3.2-3B-Instruct & 0.478 & 0.383 & 0.501 & 0.403 & 0.559 & 0.433 \\
                  & Meta-Llama-3-8B-Instruct & 0.525 & 0.372 & 0.548 & 0.365 & 0.509 & 0.314 \\
                  & gemma-2-2b-it & 0.460 & 0.299 & 0.505 & 0.330 & 0.520 & 0.393 \\
                  & gemma-2-9b-it & 0.482 & 0.444 & 0.460 & 0.440 & 0.466 & 0.414 \\
    \midrule
    truthfulQA & Llama-3.2-3B-Instruct & 0.506 & 0.311 & 0.509 & 0.314 & 0.499 & 0.305 \\
               & Meta-Llama-3-8B-Instruct & 0.550 & 0.392 & 0.540 & 0.383 & 0.538 & 0.377 \\
               & gemma-2-2b-it & 0.566 & 0.357 & 0.529 & 0.334 & 0.491 & 0.306 \\
               & gemma-2-9b-it & 0.425 & 0.496 & 0.430 & 0.509 & 0.486 & 0.526 \\
    \midrule
    gsm8k & Llama-3.2-3B-Instruct & 0.601 & 0.454 & 0.612 & 0.437 & 0.649 & 0.467 \\
          & Meta-Llama-3-8B-Instruct & 0.554 & 0.424 & 0.568 & 0.438 & 0.579 & 0.433 \\
          & gemma-2-2b-it & 0.509 & 0.398 & 0.563 & 0.394 & 0.598 & 0.395 \\
          & gemma-2-9b-it & 0.560 & 0.523 & 0.591 & 0.535 & 0.617 & 0.526 \\
    \bottomrule
  \end{tabular}
  }
\end{table}

\section{Additional Analysis Result}
\label{additional_result}

\begin{figure}[H]
    \centering
    \includegraphics[width=0.9\linewidth]{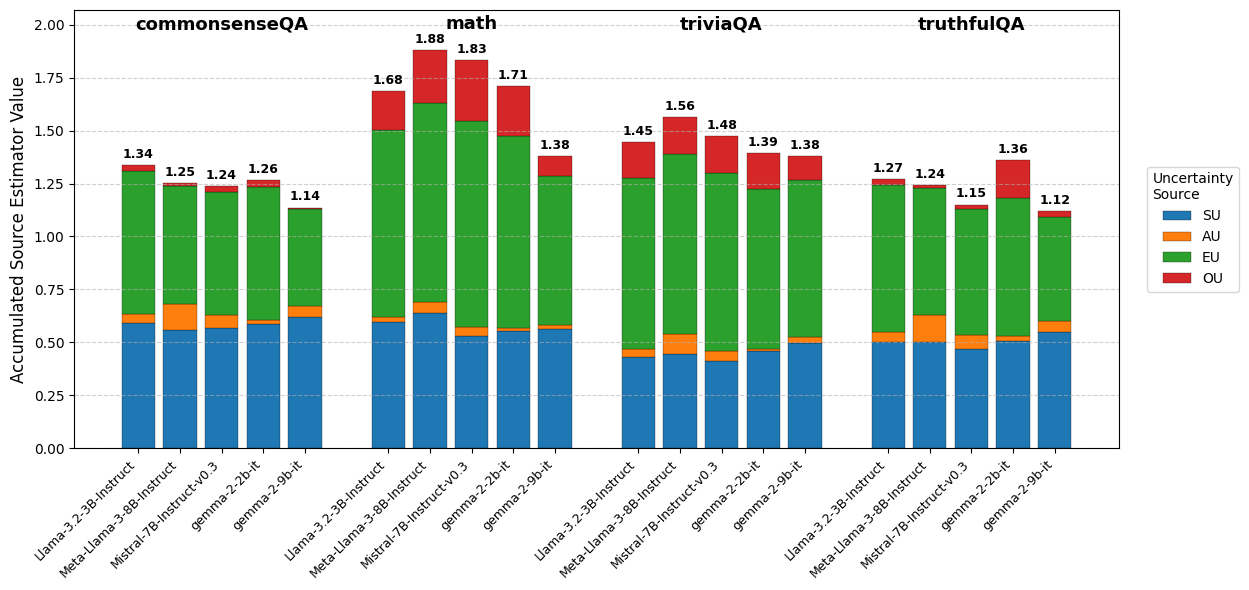}
    \caption{Uncertainty Source Decomposition of each dataset-model pair.}
    \label{fig:source_per_model_data}
\end{figure}

\begin{figure}
    \centering
    \includegraphics[width=0.9\linewidth]{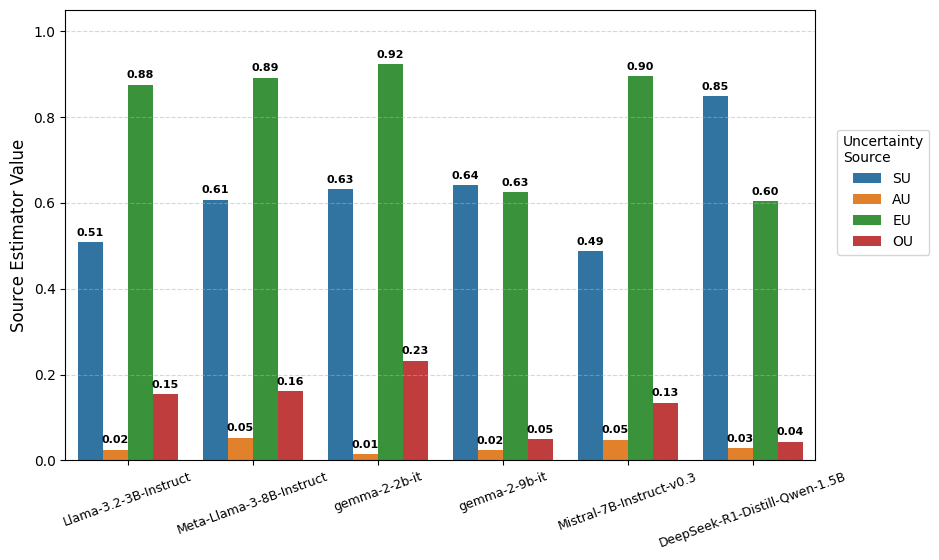}
    \caption{\textbf{Uncertainty Source Comparison: Reasoning Model vs. Other LLMs.} This comparison shows that \textsc{DeepSeek-R1-Distill-Qwen-1.5B} exhibits a higher level of SU and lower EU and OU compared to other models, highlighting its distinct uncertainty profile when tackling reasoning tasks.}
    \label{fig:rm_on_math}
\end{figure}

\begin{figure}
    \centering
    \includegraphics[width=0.9\linewidth]{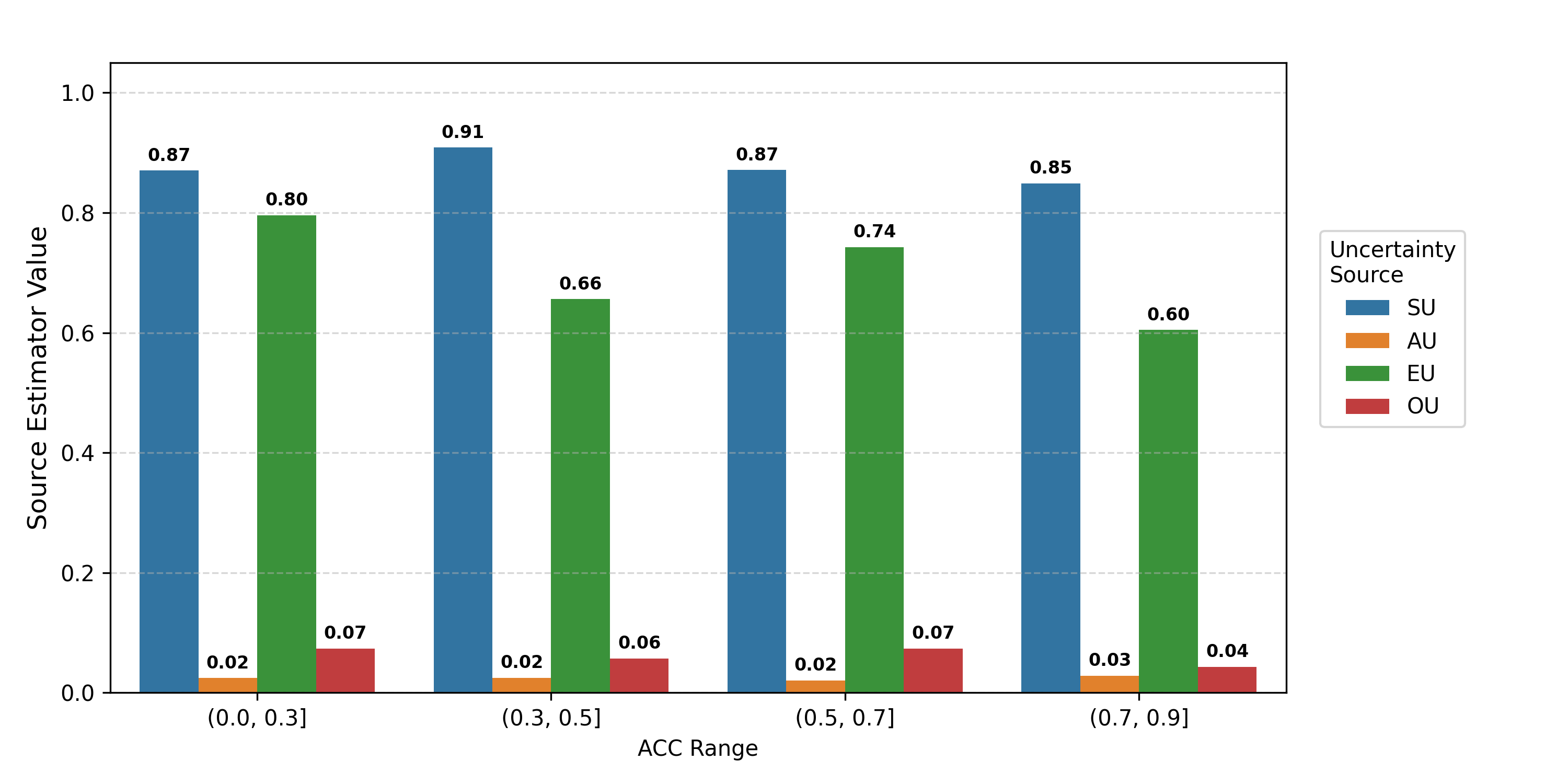}
    \caption{\textbf{DeepSeek-R1-Distill-Qwen-1.5B Uncertainty Variation Across Difficulty Levels on the MATH Dataset.} The figure shows that the uncertainty sources remain stable across varying task difficulties, unlike non-reasoning models which exhibit a downward trend in EU and OU as the difficulty increases.}
    \label{fig:rm_difficulty_variation}
\end{figure}

\section{Prompt Templates}
\label{prompt_template}

\begin{figure}[H]
    \centering
    \begin{tcolorbox}[
        width=1.0\linewidth, halign=left, 
        colframe=black, colback=white, 
        boxsep=0.01mm, arc=1.5mm, 
        left=2mm, right=2mm, boxrule=0.5pt]
    \footnotesize

    Paraphrase the following question, without changing its meaning. \\
    Make sure you only output a single question only. \\
    Question: \{q\} \\
    Paraphrased Question:

    \end{tcolorbox}
    \caption{Paraphrased Prompt Template}
    \label{fig:paraphrased_prompt}
\end{figure}

\begin{figure}[H]
    \centering
    \begin{tcolorbox}[
        width=1.0\linewidth, halign=left, 
        colframe=black, colback=white, 
        boxsep=0.01mm, arc=1.5mm, 
        left=2mm, right=2mm, boxrule=0.5pt]
    \footnotesize

    Clarify the following question by rewriting it in a clearer, more complete form.\\
    If the question is ambiguous, add missing details to make it understandable.\\
    Make sure you only output a single question only.    \\
    Original Question: \{q\} \\
    Clarified Question:

    \end{tcolorbox}
    \caption{Clarified Prompt Template}
    \label{fig:clarify_prompt}
\end{figure}

\begin{figure}[H]
    \centering
    \begin{tcolorbox}[
        width=1.0\linewidth, halign=left, 
        colframe=black, colback=white, 
        boxsep=0.01mm, arc=1.5mm, 
        left=2mm, right=2mm, boxrule=0.5pt]
    \footnotesize

    Please answer the following question. Think carefully and in a step-by-step fashion. \\
    At the end of your solution, indicate your final answer by writing the answer choice (A, B, C, D, or E) inside a boxed environment, like: \boxed{A}. \\
    Q: \{q\} \\
    Choices: \{c\} \\
    Your answer: 
    
    \end{tcolorbox}
    \caption{Sampling Prompt Template for MC Questions}
    \label{fig:sc_sampling_prompt}
\end{figure}

\begin{figure}[H]
    \centering
    \begin{tcolorbox}[
        width=1.0\linewidth, halign=left, 
        colframe=black, colback=white, 
        boxsep=0.01mm, arc=1.5mm, 
        left=2mm, right=2mm, boxrule=0.5pt]
    \footnotesize

    Following is your previous response to the question. \\
    Q: \{q\} \\
    Choices: \{c\} \\
    Your previous response: \{a\} \\

    Check your previous response carefully and solve the same question again. \\
    At the end of your solution, indicate your final answer by writing one of the answer choice (only letter : A, B, C, D, or E) inside a boxed environment, like: \boxed{A}. \\
    Output:

    \end{tcolorbox}
    \caption{Check Prompt Template for MC Questions}
    \label{fig:sc_check_prompt}
\end{figure}

\begin{figure}[H]
    \centering
    \begin{tcolorbox}[
        width=1.0\linewidth, halign=left, 
        colframe=black, colback=white, 
        boxsep=0.01mm, arc=1.5mm, 
        left=2mm, right=2mm, boxrule=0.5pt]
    \footnotesize

        Read the following passage and answer the question. \\
        Passage : \{p\} \\
        Question : \{q\} \\ 
        At the end of your solution, indicate your final answer inside a boxed environment, like: \boxed{answer}.
    
    \end{tcolorbox}
    \caption{Sampling Prompt Template for RC Questions}
    \label{fig:rc_sampling_prompt}
\end{figure}

\begin{figure}[H]
    \centering
    \begin{tcolorbox}[
        width=1.0\linewidth, halign=left, 
        colframe=black, colback=white, 
        boxsep=0.01mm, arc=1.5mm, 
        left=2mm, right=2mm, boxrule=0.5pt]
    \footnotesize

    Following is your previous response to the question: \\
    Read the following passage and answer the question. \\
    Passage : \{p\} \\
    Question : \{q\} \\ 

    Your previous response: \{a\} \\
    Check your previous response carefully and respond the question again. \\
    At the end of your solution, indicate your final answer inside a boxed environment, like: \boxed{answer}.

    \end{tcolorbox}
    \caption{Check Prompt Template for RC Questions}
    \label{fig:rc_check_prompt}
\end{figure}

\begin{figure}[H]
    \centering
    \begin{tcolorbox}[
        width=1.0\linewidth, halign=left, 
        colframe=black, colback=white, 
        boxsep=0.01mm, arc=1.5mm, 
        left=2mm, right=2mm, boxrule=0.5pt]
    \footnotesize

    Please answer the following question. \\
    Think carefully and in a step-by-step fashion. \\ 
    At the end of your solution, put your final result in a boxed environment, 
    e.g. \boxed{answer}. \\
    Q: \{q\}  \\

    \end{tcolorbox}
    \caption{Sampling Prompt Template for Essay Questions}
    \label{fig:es_sampling_prompt}
\end{figure}

\begin{figure}[H]
    \centering
    \begin{tcolorbox}[
        width=1.0\linewidth, halign=left, 
        colframe=black, colback=white, 
        boxsep=0.01mm, arc=1.5mm, 
        left=2mm, right=2mm, boxrule=0.5pt]
    \footnotesize

    Following is your previous response to the question. \\
    Q: \{q\} \\
    Your previous response: \{a\} \\
    Check your previous response carefully and solve the same question again step by step. \\
    At the end of your solution, put your final result in a boxed environment,
    eg. (\boxed{answer}). \\
    Output:

    \end{tcolorbox}
    \caption{Check Prompt Template for Essay Questions}
    \label{fig:es_check_prompt}
\end{figure}

\section{Limitations}
\label{sec:limitation}
\textbf{Estimator Validity.} While our method disentangles different sources of uncertainty, we acknowledge that the estimators may not achieve perfect separation. Uncertainty can propagate across stages of the prompting pipeline, and interactions between sources may lead to confounding effects that blur the distinctions between them. \textbf{Model Sensitivity.} We observe that models with different sizes respond differently to the same levels of estimated uncertainty. In particular, smaller models tend to be more sensitive, showing a significant performance degradation with small increases in uncertainty, whereas larger models appear to be more robust. Understanding and modeling this discrepancy remains an open direction for future work. \textbf{Evaluation Scope.} Due to computational constraints, we evaluate on a sampled subset of benchmark datasets. Although we employ bootstrapping to mitigate sampling bias, our findings may still be affected by limited scale.

\clearpage

\clearpage
\section*{NeurIPS Paper Checklist}
The checklist is designed to encourage best practices for responsible machine learning research, addressing issues of reproducibility, transparency, research ethics, and societal impact. Do not remove the checklist: {\bf The papers not including the checklist will be desk rejected.} The checklist should follow the references and follow the (optional) supplemental material.  The checklist does NOT count towards the page
limit. 

Please read the checklist guidelines carefully for information on how to answer these questions. For each question in the checklist:
\begin{itemize}
    \item You should answer \answerYes{}, \answerNo{}, or \answerNA{}.
    \item \answerNA{} means either that the question is Not Applicable for that particular paper or the relevant information is Not Available.
    \item Please provide a short (1–2 sentence) justification right after your answer (even for NA). 
\end{itemize}

{\bf The checklist answers are an integral part of your paper submission.} They are visible to the reviewers, area chairs, senior area chairs, and ethics reviewers. You will be asked to also include it (after eventual revisions) with the final version of your paper, and its final version will be published with the paper.

The reviewers of your paper will be asked to use the checklist as one of the factors in their evaluation. While "\answerYes{}" is generally preferable to "\answerNo{}", it is perfectly acceptable to answer "\answerNo{}" provided a proper justification is given (e.g., "error bars are not reported because it would be too computationally expensive" or "we were unable to find the license for the dataset we used"). In general, answering "\answerNo{}" or "\answerNA{}" is not grounds for rejection. While the questions are phrased in a binary way, we acknowledge that the true answer is often more nuanced, so please just use your best judgment and write a justification to elaborate. All supporting evidence can appear either in the main paper or the supplemental material, provided in appendix. If you answer \answerYes{} to a question, in the justification please point to the section(s) where related material for the question can be found.

IMPORTANT, please:
\begin{itemize}
    \item {\bf Delete this instruction block, but keep the section heading ``NeurIPS Paper Checklist"},
    \item  {\bf Keep the checklist subsection headings, questions/answers and guidelines below.}
    \item {\bf Do not modify the questions and only use the provided macros for your answers}.
\end{itemize}


\begin{enumerate}

\item {\bf Claims}
    \item[] Question: Do the main claims made in the abstract and introduction accurately reflect the paper's contributions and scope?
    \item[] Answer: \answerYes{} 
    \item[] Justification: The abstract and introduction accurately describe the paper’s contributions, including the uncertainty decomposition, estimation framework, and profile-guided model/metric selection. These are well-supported by the methods and experiments.
    \item[] Guidelines:
    \begin{itemize}
        \item The answer NA means that the abstract and introduction do not include the claims made in the paper.
        \item The abstract and/or introduction should clearly state the claims made, including the contributions made in the paper and important assumptions and limitations. A No or NA answer to this question will not be perceived well by the reviewers. 
        \item The claims made should match theoretical and experimental results, and reflect how much the results can be expected to generalize to other settings. 
        \item It is fine to include aspirational goals as motivation as long as it is clear that these goals are not attained by the paper. 
    \end{itemize}

\item {\bf Limitations}
    \item[] Question: Does the paper discuss the limitations of the work performed by the authors?
    \item[] Answer: \answerYes{} 
    \item[] Justification: The paper discusses limitations in Section~\ref{sec:limitation}, noting that source disentanglement may be imperfect and that computational constraints limit broader evaluations. These points help contextualize the scope and robustness of the proposed method.
    \item[] Guidelines:
    \begin{itemize}
        \item The answer NA means that the paper has no limitation while the answer No means that the paper has limitations, but those are not discussed in the paper. 
        \item The authors are encouraged to create a separate "Limitations" section in their paper.
        \item The paper should point out any strong assumptions and how robust the results are to violations of these assumptions (e.g., independence assumptions, noiseless settings, model well-specification, asymptotic approximations only holding locally). The authors should reflect on how these assumptions might be violated in practice and what the implications would be.
        \item The authors should reflect on the scope of the claims made, e.g., if the approach was only tested on a few datasets or with a few runs. In general, empirical results often depend on implicit assumptions, which should be articulated.
        \item The authors should reflect on the factors that influence the performance of the approach. For example, a facial recognition algorithm may perform poorly when image resolution is low or images are taken in low lighting. Or a speech-to-text system might not be used reliably to provide closed captions for online lectures because it fails to handle technical jargon.
        \item The authors should discuss the computational efficiency of the proposed algorithms and how they scale with dataset size.
        \item If applicable, the authors should discuss possible limitations of their approach to address problems of privacy and fairness.
        \item While the authors might fear that complete honesty about limitations might be used by reviewers as grounds for rejection, a worse outcome might be that reviewers discover limitations that aren't acknowledged in the paper. The authors should use their best judgment and recognize that individual actions in favor of transparency play an important role in developing norms that preserve the integrity of the community. Reviewers will be specifically instructed to not penalize honesty concerning limitations.
    \end{itemize}

\item {\bf Theory assumptions and proofs}
    \item[] Question: For each theoretical result, does the paper provide the full set of assumptions and a complete (and correct) proof?
    \item[] Answer: \answerNA{} 
    \item[] Justification: No theoretical results.
    \item[] Guidelines:
    \begin{itemize}
        \item The answer NA means that the paper does not include theoretical results. 
        \item All the theorems, formulas, and proofs in the paper should be numbered and cross-referenced.
        \item All assumptions should be clearly stated or referenced in the statement of any theorems.
        \item The proofs can either appear in the main paper or the supplemental material, but if they appear in the supplemental material, the authors are encouraged to provide a short proof sketch to provide intuition. 
        \item Inversely, any informal proof provided in the core of the paper should be complemented by formal proofs provided in appendix or supplemental material.
        \item Theorems and Lemmas that the proof relies upon should be properly referenced. 
    \end{itemize}

    \item {\bf Experimental result reproducibility}
    \item[] Question: Does the paper fully disclose all the information needed to reproduce the main experimental results of the paper to the extent that it affects the main claims and/or conclusions of the paper (regardless of whether the code and data are provided or not)?
    \item[] Answer: \answerYes{} 
    \item[] Justification: The paper provides detailed experimental settings in Section~\ref{sec:profile:implementation}, including datasets, models, sampling procedures, and evaluation metrics, sufficient to reproduce the main results.
    \item[] Guidelines:
    \begin{itemize}
        \item The answer NA means that the paper does not include experiments.
        \item If the paper includes experiments, a No answer to this question will not be perceived well by the reviewers: Making the paper reproducible is important, regardless of whether the code and data are provided or not.
        \item If the contribution is a dataset and/or model, the authors should describe the steps taken to make their results reproducible or verifiable. 
        \item Depending on the contribution, reproducibility can be accomplished in various ways. For example, if the contribution is a novel architecture, describing the architecture fully might suffice, or if the contribution is a specific model and empirical evaluation, it may be necessary to either make it possible for others to replicate the model with the same dataset, or provide access to the model. In general. releasing code and data is often one good way to accomplish this, but reproducibility can also be provided via detailed instructions for how to replicate the results, access to a hosted model (e.g., in the case of a large language model), releasing of a model checkpoint, or other means that are appropriate to the research performed.
        \item While NeurIPS does not require releasing code, the conference does require all submissions to provide some reasonable avenue for reproducibility, which may depend on the nature of the contribution. For example
        \begin{enumerate}
            \item If the contribution is primarily a new algorithm, the paper should make it clear how to reproduce that algorithm.
            \item If the contribution is primarily a new model architecture, the paper should describe the architecture clearly and fully.
            \item If the contribution is a new model (e.g., a large language model), then there should either be a way to access this model for reproducing the results or a way to reproduce the model (e.g., with an open-source dataset or instructions for how to construct the dataset).
            \item We recognize that reproducibility may be tricky in some cases, in which case authors are welcome to describe the particular way they provide for reproducibility. In the case of closed-source models, it may be that access to the model is limited in some way (e.g., to registered users), but it should be possible for other researchers to have some path to reproducing or verifying the results.
        \end{enumerate}
    \end{itemize}

\item {\bf Open access to data and code}
    \item[] Question: Does the paper provide open access to the data and code, with sufficient instructions to faithfully reproduce the main experimental results, as described in supplemental material?
    \item[] Answer: \answerYes{} 
    \item[] Justification: The code and data are not yet publicly released, but the authors plan to make them available upon publication.
    \item[] Guidelines:
    \begin{itemize}
        \item The answer NA means that paper does not include experiments requiring code.
        \item Please see the NeurIPS code and data submission guidelines (\url{https://nips.cc/public/guides/CodeSubmissionPolicy}) for more details.
        \item While we encourage the release of code and data, we understand that this might not be possible, so “No” is an acceptable answer. Papers cannot be rejected simply for not including code, unless this is central to the contribution (e.g., for a new open-source benchmark).
        \item The instructions should contain the exact command and environment needed to run to reproduce the results. See the NeurIPS code and data submission guidelines (\url{https://nips.cc/public/guides/CodeSubmissionPolicy}) for more details.
        \item The authors should provide instructions on data access and preparation, including how to access the raw data, preprocessed data, intermediate data, and generated data, etc.
        \item The authors should provide scripts to reproduce all experimental results for the new proposed method and baselines. If only a subset of experiments are reproducible, they should state which ones are omitted from the script and why.
        \item At submission time, to preserve anonymity, the authors should release anonymized versions (if applicable).
        \item Providing as much information as possible in supplemental material (appended to the paper) is recommended, but including URLs to data and code is permitted.
    \end{itemize}

\item {\bf Experimental setting/details}
    \item[] Question: Does the paper specify all the training and test details (e.g., data splits, hyperparameters, how they were chosen, type of optimizer, etc.) necessary to understand the results?
    \item[] Answer: \answerYes{} 
    \item[] Justification: The paper provides key experimental details in Appendix~\ref{app:implementation-detail}.
    \item[] Guidelines:
    \begin{itemize}
        \item The answer NA means that the paper does not include experiments.
        \item The experimental setting should be presented in the core of the paper to a level of detail that is necessary to appreciate the results and make sense of them.
        \item The full details can be provided either with the code, in appendix, or as supplemental material.
    \end{itemize}

\item {\bf Experiment statistical significance}
    \item[] Question: Does the paper report error bars suitably and correctly defined or other appropriate information about the statistical significance of the experiments?
    \item[] Answer: \answerYes{} 
    \item[] Justification: The paper reports statistical significance using AUROC/AUPRC metrics and applies bootstrapping for robustness, as detailed in Appendix~\ref{sec:profile:stats}.
    \item[] Guidelines:
    \begin{itemize}
        \item The answer NA means that the paper does not include experiments.
        \item The authors should answer "Yes" if the results are accompanied by error bars, confidence intervals, or statistical significance tests, at least for the experiments that support the main claims of the paper.
        \item The factors of variability that the error bars are capturing should be clearly stated (for example, train/test split, initialization, random drawing of some parameter, or overall run with given experimental conditions).
        \item The method for calculating the error bars should be explained (closed form formula, call to a library function, bootstrap, etc.)
        \item The assumptions made should be given (e.g., Normally distributed errors).
        \item It should be clear whether the error bar is the standard deviation or the standard error of the mean.
        \item It is OK to report 1-sigma error bars, but one should state it. The authors should preferably report a 2-sigma error bar than state that they have a 96\% CI, if the hypothesis of Normality of errors is not verified.
        \item For asymmetric distributions, the authors should be careful not to show in tables or figures symmetric error bars that would yield results that are out of range (e.g. negative error rates).
        \item If error bars are reported in tables or plots, The authors should explain in the text how they were calculated and reference the corresponding figures or tables in the text.
    \end{itemize}

\item {\bf Experiments compute resources}
    \item[] Question: For each experiment, does the paper provide sufficient information on the computer resources (type of compute workers, memory, time of execution) needed to reproduce the experiments?
    \item[] Answer: \answerYes{} 
    \item[] Justification: All experiments are conducted on a server equipped with two NVIDIA GeForce RTX 3090 GPUs (24GB VRAM each, CUDA Version 12.6) and dual Intel(R) Xeon(R) Gold 6230 CPUs (80 threads total, 2.10GHz). Each experiment exclusively uses a single GPU. We use the vLLM engine for efficient inference and the quantized versions of models from Hugging Face for hidden state extraction. More in Appendix~\ref{sec:profile:compute}.

    \item[] Guidelines:
    \begin{itemize}
        \item The answer NA means that the paper does not include experiments.
        \item The paper should indicate the type of compute workers CPU or GPU, internal cluster, or cloud provider, including relevant memory and storage.
        \item The paper should provide the amount of compute required for each of the individual experimental runs as well as estimate the total compute. 
        \item The paper should disclose whether the full research project required more compute than the experiments reported in the paper (e.g., preliminary or failed experiments that didn't make it into the paper). 
    \end{itemize}
    
\item {\bf Code of ethics}
    \item[] Question: Does the research conducted in the paper conform, in every respect, with the NeurIPS Code of Ethics \url{https://neurips.cc/public/EthicsGuidelines}?
    \item[] Answer: \answerYes{} 
    \item[] Justification: The research adheres to the NeurIPS Code of Ethics, with no human subjects or sensitive data involved, and the study focuses on improving reliability and interpretability in LLM evaluation.
    \item[] Guidelines:
    \begin{itemize}
        \item The answer NA means that the authors have not reviewed the NeurIPS Code of Ethics.
        \item If the authors answer No, they should explain the special circumstances that require a deviation from the Code of Ethics.
        \item The authors should make sure to preserve anonymity (e.g., if there is a special consideration due to laws or regulations in their jurisdiction).
    \end{itemize}

\item {\bf Broader impacts}
    \item[] Question: Does the paper discuss both potential positive societal impacts and negative societal impacts of the work performed?
    \item[] Answer: \answerNA{} 
    \item[] Justification: The paper focuses on methodological advances in uncertainty estimation for language models and does not directly involve societal applications, thus societal impacts are not discussed.
    \item[] Guidelines:
    \begin{itemize}
        \item The answer NA means that there is no societal impact of the work performed.
        \item If the authors answer NA or No, they should explain why their work has no societal impact or why the paper does not address societal impact.
        \item Examples of negative societal impacts include potential malicious or unintended uses (e.g., disinformation, generating fake profiles, surveillance), fairness considerations (e.g., deployment of technologies that could make decisions that unfairly impact specific groups), privacy considerations, and security considerations.
        \item The conference expects that many papers will be foundational research and not tied to particular applications, let alone deployments. However, if there is a direct path to any negative applications, the authors should point it out. For example, it is legitimate to point out that an improvement in the quality of generative models could be used to generate deepfakes for disinformation. On the other hand, it is not needed to point out that a generic algorithm for optimizing neural networks could enable people to train models that generate Deepfakes faster.
        \item The authors should consider possible harms that could arise when the technology is being used as intended and functioning correctly, harms that could arise when the technology is being used as intended but gives incorrect results, and harms following from (intentional or unintentional) misuse of the technology.
        \item If there are negative societal impacts, the authors could also discuss possible mitigation strategies (e.g., gated release of models, providing defenses in addition to attacks, mechanisms for monitoring misuse, mechanisms to monitor how a system learns from feedback over time, improving the efficiency and accessibility of ML).
    \end{itemize}
    
\item {\bf Safeguards}
    \item[] Question: Does the paper describe safeguards that have been put in place for responsible release of data or models that have a high risk for misuse (e.g., pretrained language models, image generators, or scraped datasets)?
    \item[] Answer: \answerNA{} 
    \item[] Justification: The paper does not release any models or datasets with high risk of misuse, so safeguards are not applicable.
    \item[] Guidelines:
    \begin{itemize}
        \item The answer NA means that the paper poses no such risks.
        \item Released models that have a high risk for misuse or dual-use should be released with necessary safeguards to allow for controlled use of the model, for example by requiring that users adhere to usage guidelines or restrictions to access the model or implementing safety filters. 
        \item Datasets that have been scraped from the Internet could pose safety risks. The authors should describe how they avoided releasing unsafe images.
        \item We recognize that providing effective safeguards is challenging, and many papers do not require this, but we encourage authors to take this into account and make a best faith effort.
    \end{itemize}

\item {\bf Licenses for existing assets}
    \item[] Question: Are the creators or original owners of assets (e.g., code, data, models), used in the paper, properly credited and are the license and terms of use explicitly mentioned and properly respected?
    \item[] Answer: \answerYes{} 
    \item[] Justification: The paper uses publicly available models and datasets, all of which are properly cited in the references, with usage consistent with their respective licenses.
    \item[] Guidelines:
    \begin{itemize}
        \item The answer NA means that the paper does not use existing assets.
        \item The authors should cite the original paper that produced the code package or dataset.
        \item The authors should state which version of the asset is used and, if possible, include a URL.
        \item The name of the license (e.g., CC-BY 4.0) should be included for each asset.
        \item For scraped data from a particular source (e.g., website), the copyright and terms of service of that source should be provided.
        \item If assets are released, the license, copyright information, and terms of use in the package should be provided. For popular datasets, \url{paperswithcode.com/datasets} has curated licenses for some datasets. Their licensing guide can help determine the license of a dataset.
        \item For existing datasets that are re-packaged, both the original license and the license of the derived asset (if it has changed) should be provided.
        \item If this information is not available online, the authors are encouraged to reach out to the asset's creators.
    \end{itemize}

\item {\bf New assets}
    \item[] Question: Are new assets introduced in the paper well documented and is the documentation provided alongside the assets?
    \item[] Answer: \answerNo{} 
    \item[] Justification: New assets such as the uncertainty source estimators and selection framework are introduced, but documentation and code are planned for release upon publication and are not yet available.
    \item[] Guidelines:
    \begin{itemize}
        \item The answer NA means that the paper does not release new assets.
        \item Researchers should communicate the details of the dataset/code/model as part of their submissions via structured templates. This includes details about training, license, limitations, etc. 
        \item The paper should discuss whether and how consent was obtained from people whose asset is used.
        \item At submission time, remember to anonymize your assets (if applicable). You can either create an anonymized URL or include an anonymized zip file.
    \end{itemize}

\item {\bf Crowdsourcing and research with human subjects}
    \item[] Question: For crowdsourcing experiments and research with human subjects, does the paper include the full text of instructions given to participants and screenshots, if applicable, as well as details about compensation (if any)? 
    \item[] Answer: \answerNA{} 
    \item[] Justification: The paper does not involve crowdsourcing or research with human subjects.
    \item[] Guidelines:
    \begin{itemize}
        \item The answer NA means that the paper does not involve crowdsourcing nor research with human subjects.
        \item Including this information in the supplemental material is fine, but if the main contribution of the paper involves human subjects, then as much detail as possible should be included in the main paper. 
        \item According to the NeurIPS Code of Ethics, workers involved in data collection, curation, or other labor should be paid at least the minimum wage in the country of the data collector. 
    \end{itemize}

\item {\bf Institutional review board (IRB) approvals or equivalent for research with human subjects}
    \item[] Question: Does the paper describe potential risks incurred by study participants, whether such risks were disclosed to the subjects, and whether Institutional Review Board (IRB) approvals (or an equivalent approval/review based on the requirements of your country or institution) were obtained?
    \item[] Answer: \answerNA{} 
    \item[] Justification: The paper does not involve crowdsourcing or research with human subjects as defined by IRB standards. All manual evaluations were performed by the authors or internal collaborators and do not constitute formal human subject research. Therefore, IRB approval is not applicable.
    \item[] Guidelines:
    \begin{itemize}
        \item The answer NA means that the paper does not involve crowdsourcing nor research with human subjects.
        \item Depending on the country in which research is conducted, IRB approval (or equivalent) may be required for any human subjects research. If you obtained IRB approval, you should clearly state this in the paper. 
        \item We recognize that the procedures for this may vary significantly between institutions and locations, and we expect authors to adhere to the NeurIPS Code of Ethics and the guidelines for their institution. 
        \item For initial submissions, do not include any information that would break anonymity (if applicable), such as the institution conducting the review.
    \end{itemize}

\item {\bf Declaration of LLM usage}
    \item[] Question: Does the paper describe the usage of LLMs if it is an important, original, or non-standard component of the core methods in this research? Note that if the LLM is used only for writing, editing, or formatting purposes and does not impact the core methodology, scientific rigorousness, or originality of the research, declaration is not required.
    \item[] Answer: \answerYes{} 
    \item[] Justification: This work leverages LLMs as a core component for uncertainty profiling. These usages go beyond standard applications and are integral to the proposed methodology; they are therefore clearly described in the paper.
    \item[] Guidelines:
    \begin{itemize}
        \item The answer NA means that the core method development in this research does not involve LLMs as any important, original, or non-standard components.
        \item Please refer to our LLM policy (\url{https://neurips.cc/Conferences/2025/LLM}) for what should or should not be described.
    \end{itemize}

\end{enumerate}

\end{document}